\documentclass[pdflatex,sn-mathphys-num]{sn-jnl}% Math and Physical Sciences Numbered Reference Style
\usepackage{multirow}
\usepackage{graphicx}%
\usepackage{multirow}%
\usepackage{amsmath,amssymb,amsfonts}%
\usepackage{amsthm}%

\usepackage[title]{appendix}%
\usepackage{xcolor}%
\usepackage{textcomp}%
\usepackage{manyfoot}%
\usepackage{booktabs}%
\usepackage[ruled,vlined]{algorithm2e}
\usepackage{algpseudocode}%
\usepackage{listings}%
\usepackage{CJKutf8}
\usepackage{adjustbox}
\usepackage{makecell} % 提供 \Xcline
\usepackage[normalem]{ulem}
\useunder{\uline}{\ul}{}

\definecolor{newred}{RGB}{200,100,0}

\theoremstyle{thmstyleone}%
\theoremstyle{thmstyletwo}%

\theoremstyle{thmstylethree}%

\begin{document}

\begin{CJK*}{UTF8}{gbsn}

\title[Article Title]{LadderMoE: Ladder-Side Mixture of Experts Adapters for Bronze Inscription Recognition}


% \author*[1,2]{\fnm{First} \sur{Author}}\email{iauthor@gmail.com}



% \affil*[1]{\orgdiv{Department}, \orgname{Organization}, \orgaddress{\street{Street}, \city{City}, \postcode{100190}, \state{State}, \country{Country}}}

% \affil[2]{\orgdiv{Department}, \orgname{Organization}, \orgaddress{\street{Street}, \city{City}, \postcode{10587}, \state{State}, \country{Country}}}

% \affil[3]{\orgdiv{Department}, \orgname{Organization}, \orgaddress{\street{Street}, \city{City}, \postcode{610101}, \state{State}, \country{Country}}}

\author[1]{\fnm{Rixin} \sur{Zhou}}%\email{}
\author[2]{\fnm{Peiqiang} \sur{Qiu}}%\email{}
\author[2]{\fnm{Qian} \sur{Zhang}}%\email{}
\author*[2,3]{\fnm{Chuntao} \sur{Li}}\email{lct33@jlu.edu.cn}
\author*[1,3,4]{\fnm{Xi} \sur{Yang}}\email{yangxi21@jlu.edu.cn}

% \affil*[1]{\orgdiv{Department}, \orgname{Organization}, \orgaddress{\street{Street}, \city{City}, \postcode{100190}, \state{State}, \country{Country}}}

% \affil[1]{\orgdiv{School of Artificial Intelligence}, \orgname{Jilin University}, \orgaddress{ \city{Changchun}, \postcode{130000}, \state{Jilin}, \country{China}}}

% \affil[2]{\orgdiv{School of Archaeology}, \orgname{Jilin University}, \orgaddress{ \city{Changchun}, \postcode{130000}, \state{Jilin}, \country{China}}}

% \affil[3]{\orgdiv{Key Laboratory of Ancient Chinese Script, Culture Relics and Artificial Intelligence, Jilin University},\orgaddress{ \city{Changchun}, \postcode{130000}, \state{Jilin}, \country{China}}}

% \affil[4]{\orgdiv{Engineering Research Center of Knowledge-Driven Human-Machine Intelligence, MoE}, \orgaddress{ \city{Changchun}, \postcode{130000}, \state{Jilin}, \country{China}}}

\affil[1]{\orgdiv{School of Artificial Intelligence}, \orgname{Jilin University}}

\affil[2]{\orgdiv{School of Archaeology}, \orgname{Jilin University}}

\affil[3]{\orgdiv{Key Laboratory of Ancient Chinese Script, Culture Relics and Artificial Intelligence, Jilin University}}

\affil[4]{\orgdiv{Engineering Research Center of Knowledge-Driven Human-Machine Intelligence, MoE}}


\abstract{Bronze inscriptions (BI), engraved on ritual vessels, constitute a crucial stage of early Chinese writing and provide indispensable evidence for archaeological and historical studies. However, automatic BI recognition remains difficult due to severe visual degradation, multi-domain variability across photographs, rubbings, and tracings, and an extremely long-tailed character distribution. To address these challenges, we curate a large-scale BI dataset comprising 22,454 full-page images and 198,598 annotated characters spanning 6,658 unique categories, enabling robust cross-domain evaluation. Building on this resource, we develop a two-stage detection–recognition pipeline that first localizes inscriptions and then transcribes individual characters. To handle heterogeneous domains and rare classes, we equip the pipeline with LadderMoE, which augments a pretrained CLIP encoder with ladder-style MoE adapters, enabling dynamic expert specialization and stronger robustness. Comprehensive experiments on single-character and full-page recognition tasks demonstrate that our method substantially outperforms state-of-the-art scene text recognition baselines, achieving superior accuracy across head, mid, and tail categories as well as all acquisition modalities. These results establish a strong foundation for bronze inscription recognition and downstream archaeological analysis.}

%%================================%%
%% Sample for structured abstract %%
%%================================%%

% Full-Page \yang{Is this an accurate technical definition?} 
\keywords{BI Recognition, Mixture-of-Experts, Parameter-Efficient Fine-Tuning}



\maketitle

\section{Introduction}\label{sec1}

\begin{figure*}[t]
\centering
  \includegraphics[width=1\linewidth]{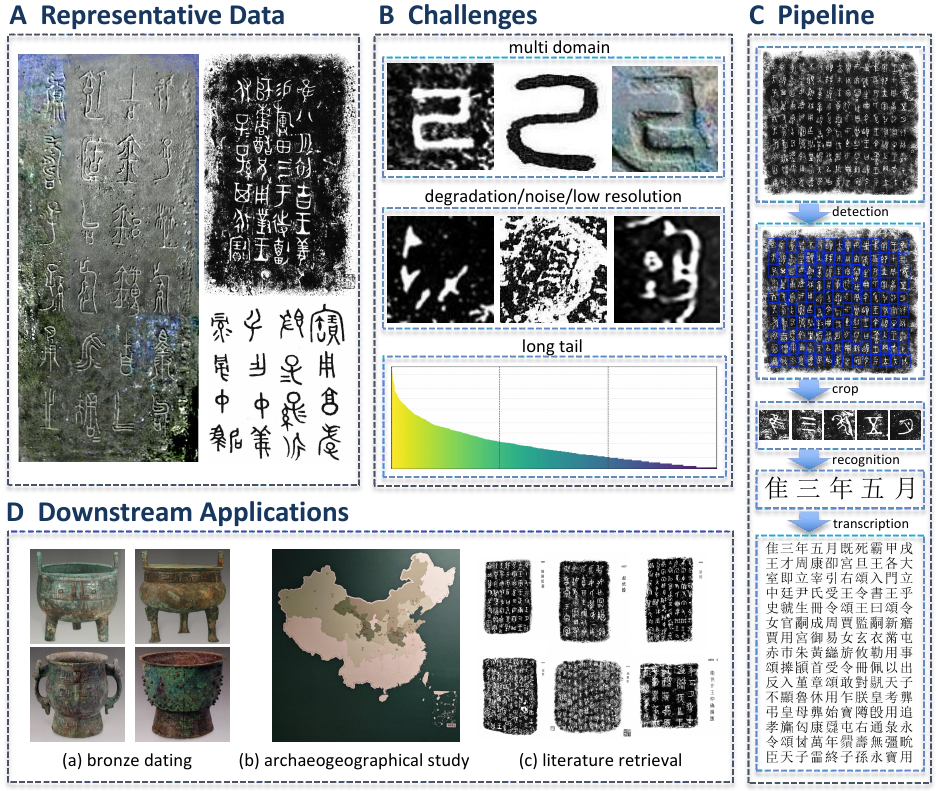}
  \caption{Overview of of our problem setting and approach for full-page bronze inscription recognition. The illustration emphasizes the cross-domain, degraded, and long-tailed nature of the data and motivates a detection–recognition–ordering framework that is robust to these factors. The resulting structured transcriptions enable downstream archaeological analyses, including bronze dating, archaeogeographical study, and literature retrieval.} 
  \label{fig: background}
\end{figure*}

Bronze inscriptions (BI), engraved on ritual vessels of ancient China, constitute a crucial component of the early Chinese writing system alongside oracle bone inscriptions (OBI), preserving invaluable records of early civilization~\cite{guo2020research}. Western Zhou inscriptions, for example, document royal rewards, sacrificial rituals, military campaigns, and political appointments~\cite{egorov2022use}. Figure~\ref{fig: background} (A) illustrates representative BI data across three typical forms: color photographs, rubbings, and tracings. Accurate recognition of such heterogeneous inscriptions is vital for downstream applications including bronze dating, archaeogeographical analysis, and historical literature retrieval, providing a reproducible bridge from raw imagery to cultural-heritage research, as shown in Figure~\ref{fig: background} (D).

Traditionally, the study of bronze inscriptions has relied on manual rubbings, tracings, and philological analysis, a process that is labor-intensive and heavily dependent on expert knowledge. With the rapid progress of computer vision, automatic detection and recognition of ancient scripts has emerged as a promising alternative. However, BI recognition remains highly challenging (Figure~\ref{fig: background}(B)) due to multi-domain diversity (color photographs, rubbings, and tracings), pronounced degradation/noise and frequent low resolution from centuries of weathering and uneven casting, and a severe long-tailed character distribution in which common ritual or administrative symbols dominate while personal names, clan titles, and toponyms are intrinsically rare~\cite{wolfgang2017language}. These factors impede the direct transfer of methods developed for OBI or modern scene text. 

Prior work has largely centered on OBI, exploring improved detectors and glyph-structure–guided methods~\cite{liu2021oracle, fu2024detecting, tao2025clustering}, with only limited extensions to BI~\cite{zheng2024ancient, wu2022cnn} that typically rely on heavy preprocessing and rubbings, leaving real-scene photographs underexplored. Meanwhile, transformer-based scene text recognition methods show promise~\cite{bautista2022scene, fang2021read, zhao2024clip4str}, but its context-aware language priors are unreliable for BI because the specialized vocabulary is scarcely represented in large pretraining corpora.

To overcome the key challenges of bronze inscription recognition and the limitations of existing research, we present the following contributions:

\begin{itemize}
\item We curate a large-scale bronze inscription dataset comprising 22,454 full-page images with 198,598 annotated character across 6,658 unique categories, spanning color photographs, rubbings, and tracings to support robust cross-domain evaluation.
\item We build a two-stage pipeline for full-page BI recognition that first detects inscriptions and then performs character recognition and transcription (Figure~\ref{fig: background} C). Within this framework, we propose the LadderMoE, a parameter-efficient model based on a pretrained CLIP image encoder that interleaves lightweight experts across multiple transformer layers, enabling efficient training and expert specialization to handle domain heterogeneity and rare-class patterns.
\item Comprehensive experiments demonstrate that our framework surpasses existing methods in overcoming the key challenges of multi-domain variation, visual degradation, and long-tailed character distribution, and achieves state-of-the-art performance on both single-character and full-page bronze inscription recognition tasks.
\end{itemize}

\section{Related Work}\label{sec2}

\subsection{Ancient Chinese Inscription Detection and Recognition}

Research on ancient Chinese script recognition has long centered on Oracle Bone Inscriptions (OBI), while Bronze Inscriptions (BI) remain comparatively underexplored. Early detection studies simply adapted generic object detectors. For example, Liu et al. enhanced Faster R-CNN for OBI character detection~\cite{liu2021oracle}, Fu et al. introduced pseudo-category labels and glyph-structure priors to improve noise robustness~\cite{fu2024detecting}, and Tao et al. leveraged the OBC font library with clustering-based representation learning for stronger feature extraction~\cite{tao2025clustering}. These methods established a foundation but still inherit the limitations of generic detectors, including heavy dependence on preprocessing and insufficient adaptability to heterogeneous visual domains. Recognition techniques have progressed along two principal directions. Structure-driven pipelines extract line or stroke-level geometry and then perform geometric matching, e.g., via Hough transforms~\cite{meng2017recognition}. Such approaches explicitly encode stroke topology but are sensitive to background clutter and low-contrast corrosion. Learning-based models embed character images into discriminative feature spaces for nearest-neighbor or sequence matching~\cite{2018A, zhang2019oracle}, and recent transformer variants—such as the improved Swin-Transformer~\cite{zheng2024ancient} with pruning-based acceleration~\cite{JGDJ202016029}—have further advanced recognition across OBI, BI, and stone engravings.. Despite these advances, systematic BI detection and recognition remain largely unexplored. Existing pipelines generally assume well-preprocessed rubbings or clean tracings, which limits their robustness to real-scene photographs that exhibit complex casting textures, multi-domain variation (color photos, rubbings, and tracings), and significant visual degradation. Furthermore, the intrinsically long-tailed character distribution in bronze inscriptions poses challenges for balanced learning and evaluation.

\subsection{Scene Text Recognition}

Scene Text Recognition (STR) aims to read text from cropped regions in natural images and has enabled applications such as understanding road signs, product labels, and document analysis~\cite{bautista2022scene}. Unlike conventional OCR, STR must handle heterogeneous fonts, arbitrary orientations, curved layouts, and complex illumination, making it a particularly challenging problem. Recent progress has been driven by transformer-based sequence models~\cite{bautista2022scene,atienza2021vision} and semi-supervised paradigms that exploit unlabeled data~\cite{aberdam2021sequence,luo2022siman}, complementing earlier end-to-end architectures~\cite{Baek_Kim_Lee_Park_Han_Yun_Oh_Lee_2019,Baek_Matsui_Aizawa_2021,bhunia2021towards}. Methodologically, STR approaches fall into two categories. Context-free methods rely solely on visual evidence, including CTC-based recognizers~\cite{graves2006connectionist,he2016reading,shi2016end,borisyuk2018rosetta}, segmentation-driven pipelines~\cite{liao2019scene,wan2020textscanner}, and attention-based encoder–decoder models~\cite{cheng2017focusing,shi2018aster}. Context-aware methods augment vision with linguistic priors, as in ABINet~\cite{fang2021read}, CLIP-OCR~\cite{wang2023symmetrical}, and CLIP4STR~\cite{zhao2024clip4str}, which leverage external language models or cross-modal knowledge. The challenges faced in bronze inscription recognition closely parallel those of STR: characters appear on complex, uneven surfaces with variable lighting, occlusion, and background noise. Context-free STR techniques—which focus purely on robust visual modeling—provide a suitable foundation for recognizing BI from both rubbings and real-scene photographs.

\subsection{Parameter-efficient Fine-tuning}

Parameter-efficient fine-tuning (PEFT) adapts large pre-trained models to downstream tasks by updating only a small subset of parameters, thereby avoiding the computational and energy costs of full fine-tuning \cite{wang2025parameter}. Representative PEFT families differ in where and what they tune: adapter tuning inserts lightweight bottleneck modules into Transformer layers \cite{zhang2023adaptive, li2021prefix}; LoRA injects trainable low-rank matrices into frozen weight paths \cite{hu2022lora}; and prompt tuning optimizes task-specific, learnable prompts while keeping backbone weights fixed \cite{lester2021power}. Orthogonal to PEFT, Mixture-of-Experts (MoE) architectures expand model capacity via multiple experts and a routing network that sparsely activates only a small subset per input, enabling near-constant per-token compute while scaling representational power \cite{jacobs1991adaptive, NEURIPS2024_f77012b4, zhou2022mixture, jiang2024mixtral}. Although prior work has extensively studied PEFT and MoE in isolation, their combination is particularly appealing for domains with strong intra-class variability and modality/style heterogeneity—such as ancient script recognition—where efficient specialization and targeted parameterization are both desirable. Our work situates itself at this intersection by introducing ladder-side MoE-Adapters, which attach adapter experts along the backbone and employ routing to learn complementary representations for different types of BI. This design couples PEFT’s low-overhead adaptation with MoE’s selective expert allocation, yielding a parameter-efficient yet specialization-aware approach to bronze inscription recognition.

\section{Methodology}\label{sec3}

\subsection{Full-page Bronze Inscriptions Recognition Pipeline}
\begin{figure*}[t]
\centering
  \includegraphics[width=1\linewidth]{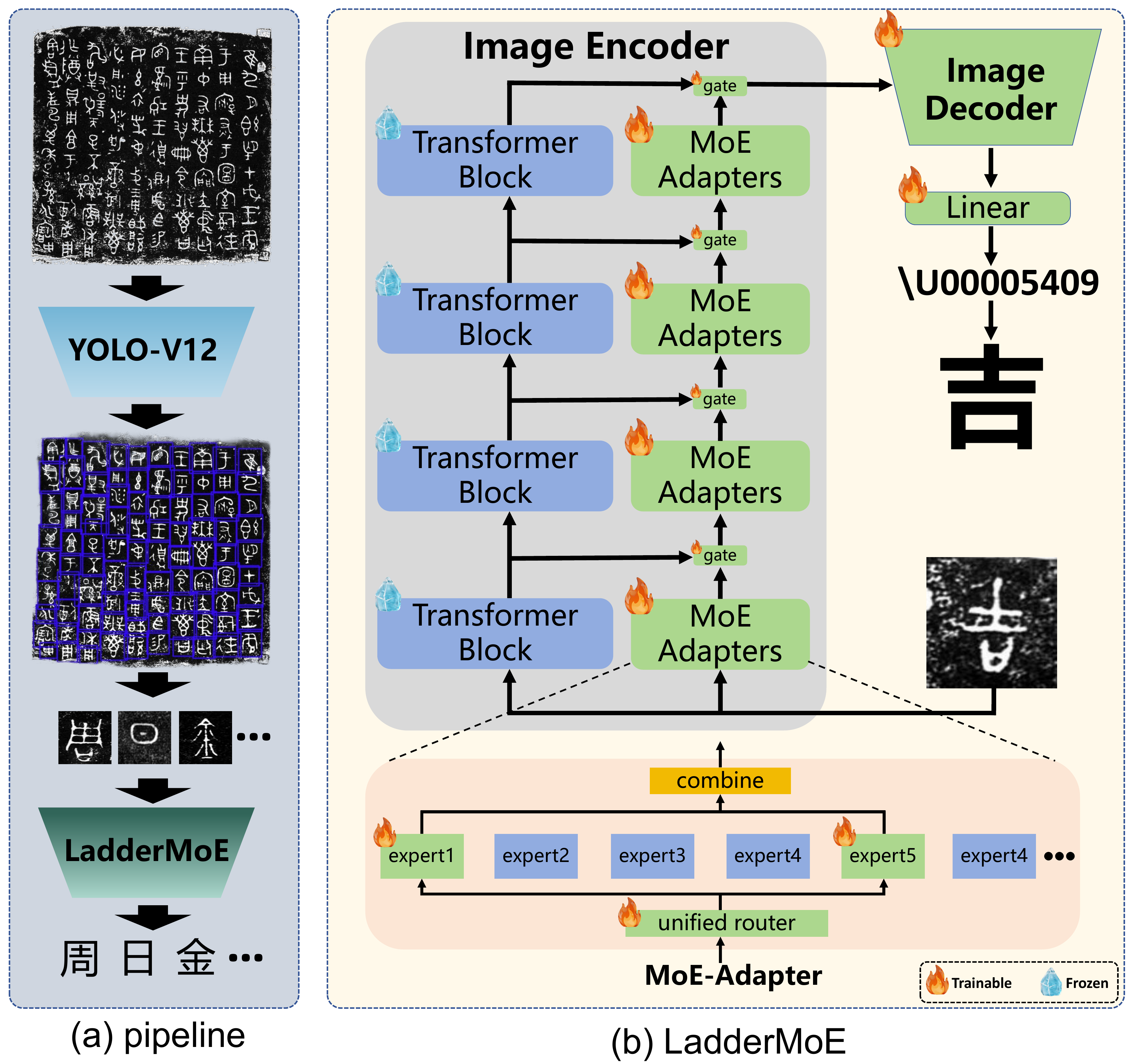}
  \caption{Framework of our bronze inscriptions recognition model. A Transformer-based encoder is augmented with interleaved MoE-Adapters, and the enriched features are decoded into character code. Each MoE-Adapter consists of multiple experts and a unified router that dynamically selects a sparse subset of experts for the input.} 
  \label{fig: network}
\end{figure*}

We adopt a two-stage detect–then–recognize pipeline for full-page BI recognition, as shown in Figure~\ref{fig: network} (a). An off-the-shelf object detector, YOLO-v12~\cite{tian2025yolov12}, is first applied to full-page inscription images to localize character instances. The detected regions are then cropped into single-character patches and recognized by our LadderMoE. During training, the detector is learned on full-page images with bounding-box annotations, and the recognizer is trained on single-character crops generated from ground-truth boxes. 

% \yang{This algorithm appears inexplicably, why？for what? Why is there such a complicated algorithm in a place where it should be an overview?}

\subsection{LadderMoE}

\subsubsection{Encoder}
As illustrated in Figure~\ref{fig: network} (b), we employ a pretrained CLIP image encoder, and MoE-Adapters are inserted at multiple intermediate layers through ladder-style connections. These adapters are governed by a unified router that dynamically selects a sparse subset of experts, enabling adaptive routing of features across categories with diverse characteristics. The outputs from the selected experts are combined and progressively fused with the backbone stream by a trainable gate, which is subsequently fed into an image decoder for final character code prediction.

\subsubsection{Ladder-side MoE Adapter}
Each MoE adapter contains a unified router responsible for selecting a sparse subset of experts from a pool of N candidate experts. Given an adapter input, the router first aggregates information from the class token and the average-pooled image token to form its routing signal. This signal is projected into a one-dimensional vector of expert scores, after which only the top-k experts with the highest scores are activated. The router then applies a softmax function to these selected scores to obtain normalized routing weights. Using these weights, the adapter computes a weighted sum of the outputs of the chosen experts, producing the final expert-enhanced representation.

\subsubsection{Decoder}

We adopt the same decoder architecture as PARSeq~\cite{bautista2022scene}, which employs a shallow single-layer decoder to extract character information from the visual feature. Unlike PARSeq, which relies on Permutation Language Modeling (PLM) for training, we further introduce an Ordered Sequence Fine-tuning (OSF) stage. The character order of BI carries intrinsic semantic meaning. Therefore, during the later phase of training we replace the random attention masks used in PLM with a fixed sequential mask. The OSF stage strengthens the alignment between the predicted character sequence and its underlying semantic structure.

\section{Datasets}\label{sec4}

\begin{figure*}[t]
\centering
  \includegraphics[width=1\linewidth]{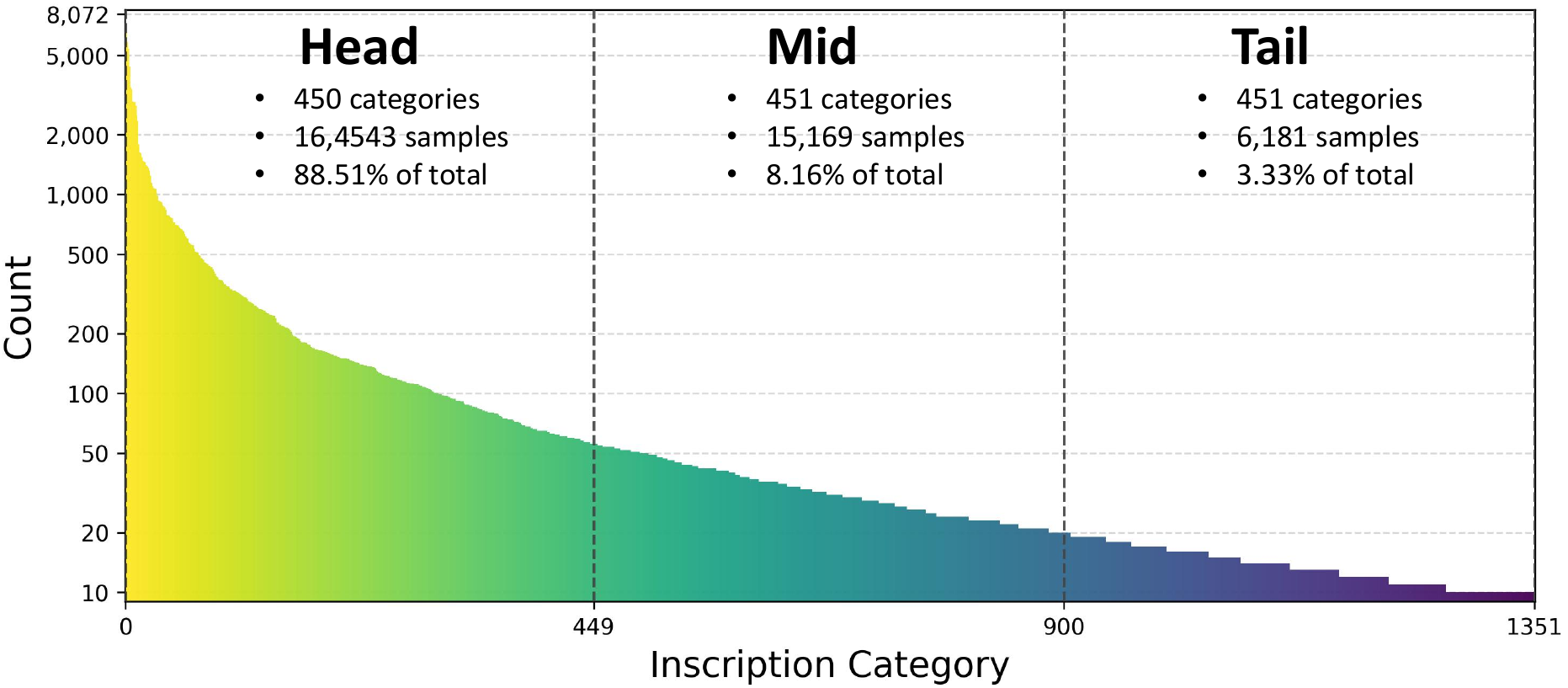}
  \caption{Category frequency distribution of the 1,352 selected bronze inscription categories. The distribution exhibits a pronounced long-tailed pattern: a few characters occur thousands of times (up to over 8,000 instances), while the majority of classes appear only sparsely. This imbalance poses a significant challenge for recognition models and motivates our evaluation across Head, Mid, and Tail subsets.} 
  \label{fig: DataDistribution}
\end{figure*}
% \yang{Is the curve in this figure correct?}

% \subsection{Datasets}
\subsection{Data Filtering} We construct a large-scale dataset of BI comprising 22,454 images covering 6,658 distinct character categories. The most frequent character appears 8,072 times, whereas some characters occur only once, reflecting the extreme sparsity. Among the collected data, 3,037 are color photographs, while the remaining 17,360 are rubbings and 2070 are tracings, capturing both archaeological records and traditional research materials. To ensure that the detection and recognition networks are trained on categories with sufficient visual evidence, we retain only those categories with more than 10 samples, discarding extremely sparse categories. The final filtered dataset contains 17,002 images across 1,352 inscription categories, and all subsequent bronze-inscription detection and recognition experiments are conducted on this refined subset.

\subsection{Data Splits}
For the full-page inscription detection and recognition task, we further split 17,002 filtered images into training, validation, and test sets with an 8 : 1 : 1 ratio, ensuring that each split preserves the same distribution of color, rubbing, and tracing images to maintain domain consistency across splits. 

To evaluate the single-char recognition task alone, we crop individual characters of the 1,352 categories from the original images, resulting in 185,893 character patches. Each character category is then divided into training, validation, and test subsets with a 4 : 1 : 5 ratio, guaranteeing balanced coverage of every category.

To enable a more comprehensive evaluation of recognition methods under category imbalance, we divide the 1,352 character categories into three groups—Head, Mid, and Tail—based on frequency, ensuring that each group contains approximately one-third of the categories, as shown in Figure~\ref{fig: DataDistribution}. This stratification allows us to analyze model performance across characters with abundant, moderate, and scarce training examples, providing insights into robustness under real-world long-tailed scenarios.

% 我们总共收集了22454张整幅金文图像，其中包含6658种正编金文字符，数量最多的字符有8072个，最少的只有1个。其中包括3046张彩色青铜器铭文照片，19408张拓片和摹本图像。训练金文检测模型的时候，我们把22454张图像按照8：1：1的比例划分成训练集，验证集，测试集。训练金文识别模型的时候，我们筛选出数量在10-8072之间的1352种金文字符，按照4：1：5划分成训练集，验证集，测试集。1352种金文字符的类别分布如图~\ref{fig: DataDistribution}所示。可以看到数量呈现严重的长尾分布。为了更全面地评估不同金文识别方法的在长尾分布的数据上的表现，我们把1352类金文字符按照数量从多到少均匀分成三份（Head， Mid， Tail），评估模型在不同数量训练样本上的表现。

\section{Experiments}\label{sec5}
\subsection{Implementation Details}
\subsubsection{Devices and Code.} 
All experiments were implemented by PyTorch, and conducted on a server with 4 RTX A40 GPUs and Intel$\circledR$ Xeon$\circledR$ Gold 5220 CPUs (72 cores). For fair comparison, we adopt the official implementations of all baseline methods.

\subsubsection{Training Details.} 
We set the batch size to 32 and train models for 40 epochs in total. Specifically, the first 35 epochs use permuted sequence masks to encourage diverse dependency learning, followed by 5 epochs ordered sequence fine-tuning, and the number of permutations for sequence modeling is set to 12. In the MoE modules, we use 36 experts per layer with top-5 expert selection.

To reduce training cost, MoE-Adapters are placed only at selected encoder layers [0, 4, 8, 11]. During training, the backbone encoder parameters are frozen, while the learnable gate, unified router, activated experts and the decoder remain learnable.

\subsubsection{Transcription Algorithm.} 
At inference, the detected boxes are first passed to the recognition model to obtain character predictions, after which Algorithm~\ref{alg:group-columns} adaptively estimates a horizontal threshold and clusters the boxes into right-to-left columns with top-to-bottom ordering, producing a structured full-page transcription result.

\begin{algorithm}[t]
\caption{Column-wise Grouping for Full-page Transcription}
\label{alg:group-columns}
\KwIn{
Detected text boxes $\mathcal{B}=\{b_i\}$, each $b=((x_1,y_1),(x_2,y_2))$; 
scaling factor $\mathrm{\lambda}$ (default $0.5$).
}
\KwOut{
Ordered columns $\mathcal{C}=[C_1,\dots,C_M]$.
}
\BlankLine
\SetKwFunction{Compute}{ComputeAdaptiveThreshold}
\SetKwProg{Fn}{Function}{:}{}
\Fn{\Compute{$\mathcal{B}$, factor}}{
    $W \leftarrow \{\,x_2 - x_1 \mid b=((x_1,y_1),(x_2,y_2))\in\mathcal{B},\; x_2>x_1 \,\}$\;
    % \uIf{$W$ is empty}{\Return $20$\tcp*[r]{default if no valid width}}
    $\overline{w} \leftarrow \dfrac{1}{|W|}\sum_{w\in W} w$\;
    \Return $\overline{w}\times \mathrm{\lambda}$\;
}
\BlankLine
\textbf{Main Procedure:}\;
$x_{\mathrm{thr}} \leftarrow$ \Compute{$\mathcal{B}$, factor}\;
Sort $\mathcal{B}$ by $x_1$ in \emph{descending} order (rightmost first)\;
$\mathcal{C} \leftarrow [\ ]$\;
\ForEach{$b\in\mathcal{B}$}{
    assigned $\leftarrow$ \textbf{false}\;
    \ForEach{$C\in\mathcal{C}$}{
        $x^{\text{anchor}} \leftarrow$ $x_1$ of the first box in $C$\;
        \If{$|x_1(b)-x^{\text{anchor}}| < x_{\mathrm{thr}}$}{
            append $b$ to $C$; assigned $\leftarrow$ \textbf{true}; \textbf{break}\;
        }
    }
    \If{\textbf{not} assigned}{
        create new column $C^\star \leftarrow [b]$ and append to $\mathcal{C}$\;
    }
}
\ForEach{$C\in\mathcal{C}$}{sort $C$ by $y_1$ in ascending order (top $\to$ bottom)}
Sort $\mathcal{C}$ by $x_1$ of the first box in \emph{descending} order (right $\to$ left)\;
\Return $\mathcal{C}$\;
\end{algorithm}

\subsubsection{Evaluation Metric.} 

\noindent\textbf{Single-Character Recognition.} We evaluate single‐inscription recognition using multiple accuracy measures to assess overall performance and robustness across class imbalance and domain shifts. Let the test set be $\mathcal{D} = \{ (x_i, y_i) \}_{i=1}^{N}$, where $x_i$ is the input and $y_i$ the ground-truth label. Denote the set of all classes as with $\mathcal{C}$ cardinality $\lvert \mathcal{C} \rvert$. For each class $c \in \mathcal{C}$, let $\mathcal{D}_c = \{ i \mid y_i = c \}$ be the index set of its samples, and let $\hat{y}_i$ be the predicted label for sample i. The indicator function $\mathbf{1}[\cdot]$ equals 1 if the condition inside is true and 0 otherwise. The overall accuracy is defined as:
\begin{equation}
\mathrm{Overall \ Acc} = \frac{1}{N} \sum_{i=1}^{N} \mathbf{1}\!\left[\hat{y}_i = y_i\right], 
\end{equation}
The class-balanced average accuracy is defined as:
\begin{equation}
\mathrm{Balanced \ Acc}
= \frac{1}{|\mathcal{C}|}
  \sum_{c \in \mathcal{C}}
  \frac{1}{|\mathcal{D}_c|}
  \sum_{i \in \mathcal{D}_c}
  \mathbf{1}\!\left[\hat{y}_i = y_i\right].
\end{equation}

To evaluate robustness across class-frequency regimes and acquisition domains, we further report accuracies on specific subsets of the test data. Let $\mathcal{D}_{\mathrm{H}},\ \mathcal{D}_{\mathrm{M}},\ \text{and}\ \mathcal{D}_{\mathrm{T}}$ denote the sample indices belonging to head, mid, and tail classes respectively. Similarly, let $\mathcal{D}_{\mathrm{d}}$ represent samples from a particular domain d (e.g., color, rubbing and tracing images). The accuracy on any subset $S \subseteq \mathcal{D}$ is defined as:
\begin{equation}
\mathrm{Subset \ Acc} = \frac{1}{|S|} \sum_{i \in S} \mathbf{1}\bigl[ \hat{y}_i = y_i \bigr].
\end{equation}

\noindent\textbf{Full-page Inscription Detection.} We evaluate the full-page BI detection performance using the standard Average Precision at a 0.5 IoU threshold ($AP_{50}$).

\noindent\textbf{Full-page Inscription Recognition.} For each page i, we serialize predicted and ground-truth character boxes into sequences $\hat{l}_i \text{ and } l_i$ using a column-first reading order (columns right-to-left; within-column top-to-bottom), then align $\hat{l}_i \text{ to } l_i$ via unit-cost Levenshtein to obtain substitution, deletion, and insertion counts $(S_i, D_i, I_i)$ and the reference length $N_i = \lvert \mathbf{l}_i \rvert$. Per-page metrics Correct Rate (CR) and Accurate Rate (AR) are defined as:
\begin{equation}
\mathrm{CR}_i=\frac{N_i-S_i-D_i}{N_i}
\end{equation}
\begin{equation}
\mathrm{AR}_i=1-\frac{S_i+D_i+I_i}{N_i}.
\end{equation}

For a dataset with M pages, we report macro and micro variants: 
\begin{equation}
\text{Macro-CR}=\frac{1}{M}\sum_{i=1}^{M}\mathrm{CR}_i, \text{Macro-AR}=\frac{1}{M}\sum_{i=1}^{M}\mathrm{AR}_i, 
\end{equation}

\begin{equation}
\text{Micro-CR}=1-\frac{\sum_{i=1}^{M}(S_i+D_i)}{\sum_{i=1}^{M}N_i}, \text{Micro-AR}=1-\frac{\sum_{i=1}^{M}(S_i+D_i+I_i)}{\sum_{i=1}^{M}N_i}.
\end{equation}

\subsection{Single-Character Recognition}

\begin{table*}[t]
\centering
\caption{Comparison of multiple evaluation settings on the single character recognition task. Our method consistently outperforms existing baselines across overall, head/mid/tail, and cross-domain (color, rubbing and tracing) evaluations. Bold numbers denote the best results and underline indicates suboptimal results.}
\label{tab:Comparison}

\resizebox{\textwidth}{!}{
\begin{tabular}{c||c|c||c|c|c||c|c|c}
\hline\hline
         & Overall Acc    & Balanced Acc   & Head Acc       & Mid Acc        & Tail Acc       & Color Acc    & Rubbing Acc  & Tracing Acc  \\ \hline\hline
ABINet~\cite{fang2021read}   & 63.64          & 18.93          & 71.05          & 9.90           & 1.63           & 50.22          & 65.98     &62.11      \\ \hline
PARSeq~\cite{bautista2022scene}   & 60.92          & 14.32          & 68.61          & 3.74           & 0.19           & 50.20          & 62.99    &58.28        \\ \hline
CLIP-OCR~\cite{wang2023symmetrical} & 67.96          & 25.35          & 74.80          & 18.58          & 5.08           & 55.14          & 69.83     &69.31     \\ \hline
CLIP4STR~\cite{zhao2024clip4str} & {\ul 76.29}    & {\ul 42.38}    & {\ul 81.79}    & {\ul 40.41}    & \textbf{20.68} & {\ul 66.80}    & {\ul 77.48}   &{\ul 78.86} \\ \hline
Ours     & \textbf{78.79} & \textbf{43.23} & \textbf{84.51} & \textbf{41.74} & {\ul 20.31}    & \textbf{70.11} & \textbf{79.96}  &\textbf{80.43} \\ \hline
\end{tabular}
}
\end{table*}

\begin{figure*}[t]
\centering
  \includegraphics[width=1\linewidth]{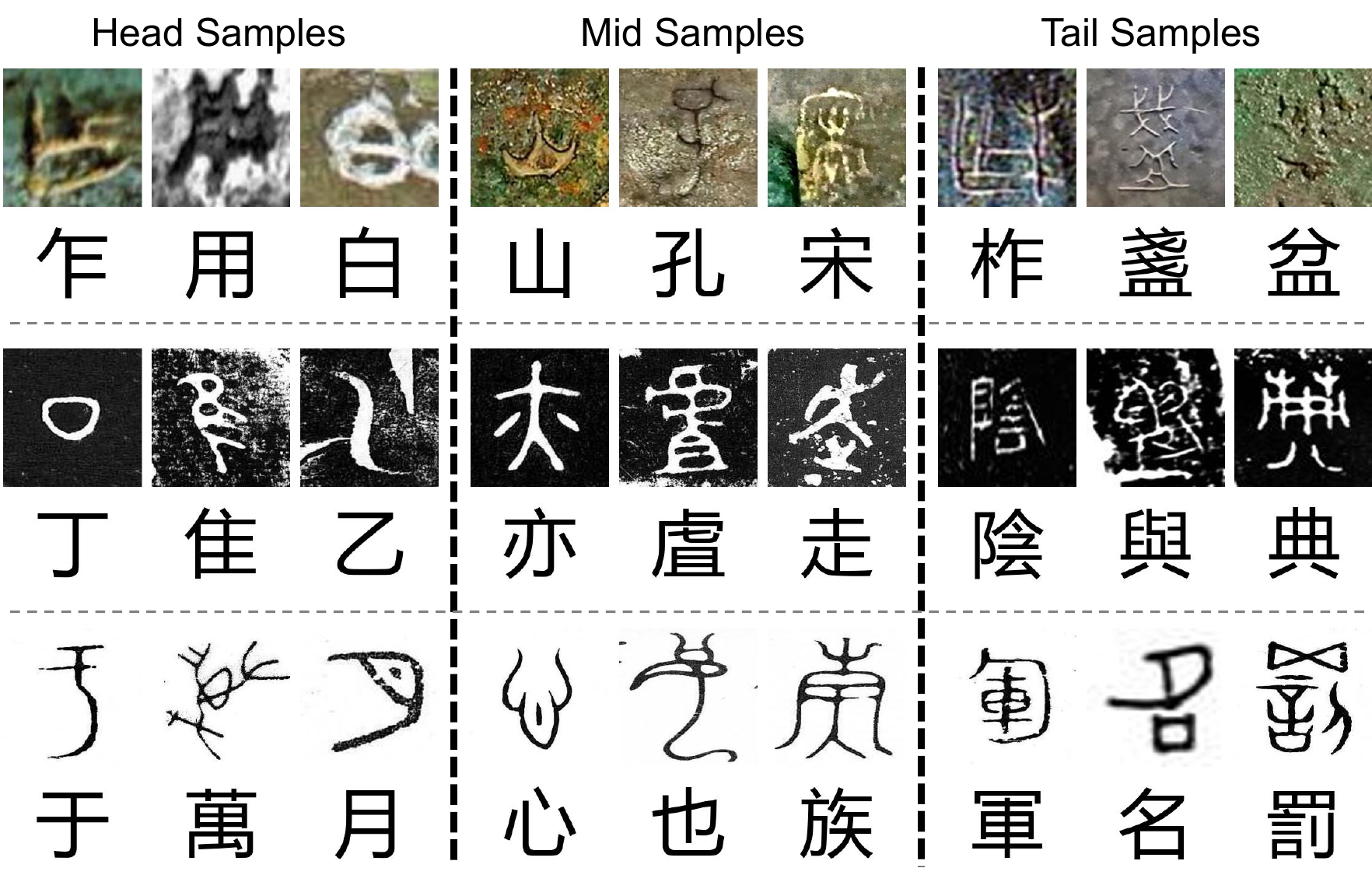}
  \caption{Correct recognition examples across frequency groups and domains. The recognition results highlight our model’s robustness to both distribution shifts and domain variations.} 
  \label{fig: recog_result_grid}
\end{figure*}

We compare our method with several representative scene text recognition approaches, as summarized in Table~\ref{tab:Comparison}. Our model achieves the best results on seven of the eight reported metrics, including an Overall Accuracy of \textbf{78.79\%} and a Balanced Accuracy of \textbf{43.23\%}, surpassing the previous best (CLIP4STR) by\% 2.5 and 0.85\%, respectively. 
For the long-tail evaluation, it reaches \textbf{84.51\%} on head classes and \textbf{41.74\%} on mid classes, and remains highly competitive on tail classes with \textbf{20.31\%}, ranking first in the former two and second in the latter. Across imaging domains, our method consistently delivers superior accuracy with \textbf{70.11\%} on color images, \textbf{79.96\%} on rubbings, and \textbf{80.43\%} on tracings. These results highlight the strong robustness of our approach under class imbalance and diverse visual domains, establishing state-of-the-art performance for single-inscription recognition.

Figure~\ref{fig: recog_result_grid} presents correctly recognized character samples across head, mid, and tail frequency groups under diverse imaging conditions. The examples show that our model accurately recognizes common characters as well as mid- and low-frequency characters that often appear with severe corrosion, low contrast, or complex textures. Notably, even tail-class samples—where training data are extremely limited and visual patterns are highly degraded—are correctly identified, underscoring the model’s strong generalization ability to rare categories and challenging acquisition domains.

\subsection{Full-page Detection and Recognition}

\begin{figure*}[t]
\centering
  \includegraphics[width=1\linewidth]{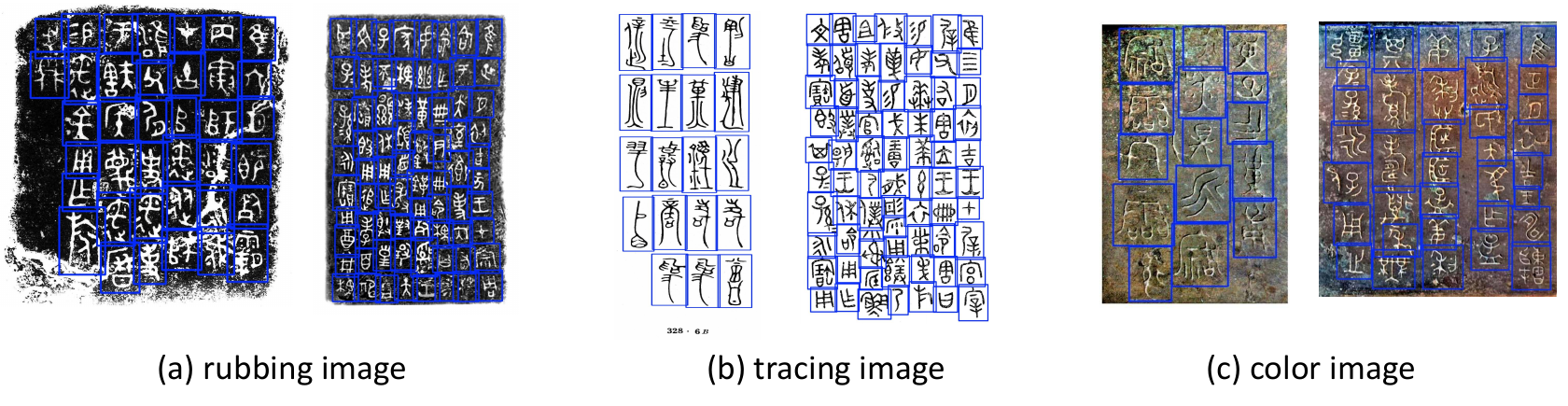}
  \caption{Detection results of YOLO-v12 on three types of inscription images: (a) rubbing images, (b) tracing images, and (c) color images. The blue bounding boxes denote the detected inscription regions.} 
  \label{fig: detection result}
\end{figure*}

\begin{table*}[t]
\centering
\caption{Performance of full-page bronze inscription detection and recognition. We report the detection metric $AP_{50}$ along with recognition metrics Macro/Micro AR and CR. Bold numbers denote the best results.}
\label{tab:full-page-comparison}
\fontsize{6pt}{8.5pt}\selectfont  % 字号=7.5pt，基线间距=8.5pt
  \setlength{\tabcolsep}{3pt}
  \renewcommand{\arraystretch}{1.03}   % 行距再略收紧一点
\begin{adjustbox}{max width=\textwidth}

\begin{tabular}{@{}c|c|c|c|c|c|c@{}}
\hline\hline
\makecell{Detection\\Method} & $AP_{50}$ & \makecell{Recognition\\Method} &
Macro-AR & Macro-CR & Micro-AR & Micro-CR \\ \hline\hline
\multirow{5}{*}{\strut YOLO-v12~\cite{tian2025yolov12}} & \multirow{5}{*}{0.8987} & ABINet~\cite{fang2021read}   & 43.45 & 65.69 & 54.05 & 67.32 \\
\Xcline{3-7}{\arrayrulewidth}
& & PARSeq~\cite{bautista2022scene}   & 40.26 & 62.54 & 49.69 & 62.96 \\
\Xcline{3-7}{\arrayrulewidth}
& & CLIP-OCR~\cite{wang2023symmetrical} & 46.47 & 68.80 & 57.23 & 70.61 \\
\Xcline{3-7}{\arrayrulewidth}
& & CLIP4STR~\cite{zhao2024clip4str} & 48.25 & 70.63 & 59.15 & 72.59 \\
\Xcline{3-7}{\arrayrulewidth}
& & Ours     & \textbf{49.67} & \textbf{72.05} & \textbf{60.10} & \textbf{73.51} \\ \hline
\end{tabular}
\end{adjustbox}
\end{table*}

We develop a complete full-page bronze inscription (BI) pipeline that first detects inscriptions and then performs end-to-end recognition. 
For detection, the YOLO-v12 model achieves an $AP_{50}$ of 0.8987, demonstrating strong capability in localizing BI instances despite complex backgrounds and diverse imaging domains. 
As shown in Figure~\ref{fig: detection result}, the model accurately highlights each inscription with bounding boxes across varied modalities and challenging textures.

Building on this detector, we integrate YOLO-v12 with multiple scene text recognition networks to construct the full-page BI recognition pipeline. 
Table~\ref{tab:full-page-comparison} reports the best performance of our pipeline, reaching 49.67\% Macro-AR, 72.05\% Macro-CR, 60.10\% Micro-AR, and 73.51\% Micro-CR. These results confirm that the recognition module not only achieves high single-character accuracy but also scales effectively to the full-page setting, validating the robustness of the overall detection–recognition framework.

\subsection{Ablation Studies}

\begin{figure*}[t]
\centering
  \includegraphics[width=1\linewidth]{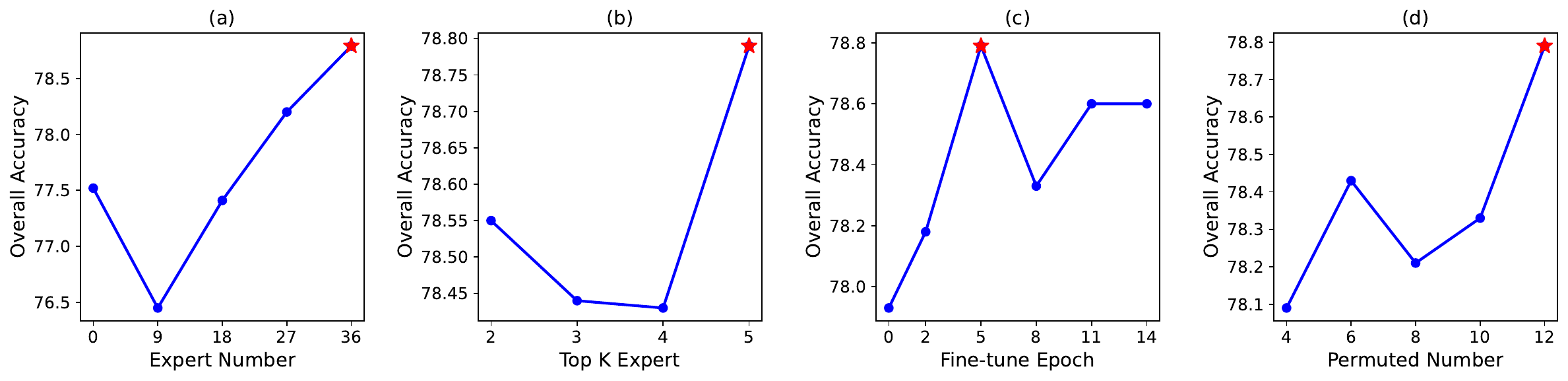}
  \caption{Overall accuracy versus (a) Varying the number of experts per MoE-Adapter, (b) top-k expert routing, (c) OSF epochs, and (d) PLM permutation count. Red stars denote the optimal settings (36 experts, top-5 routing, 5 OSF epochs, 12 permutations) achieving 78.8\% overall accuracy. Expert number, top-k routing, and permutation count show a clear upward correlation with performance.} 
  \label{fig: ablation}
\end{figure*}

\begin{figure*}[t]
\centering
  \includegraphics[width=1\linewidth]{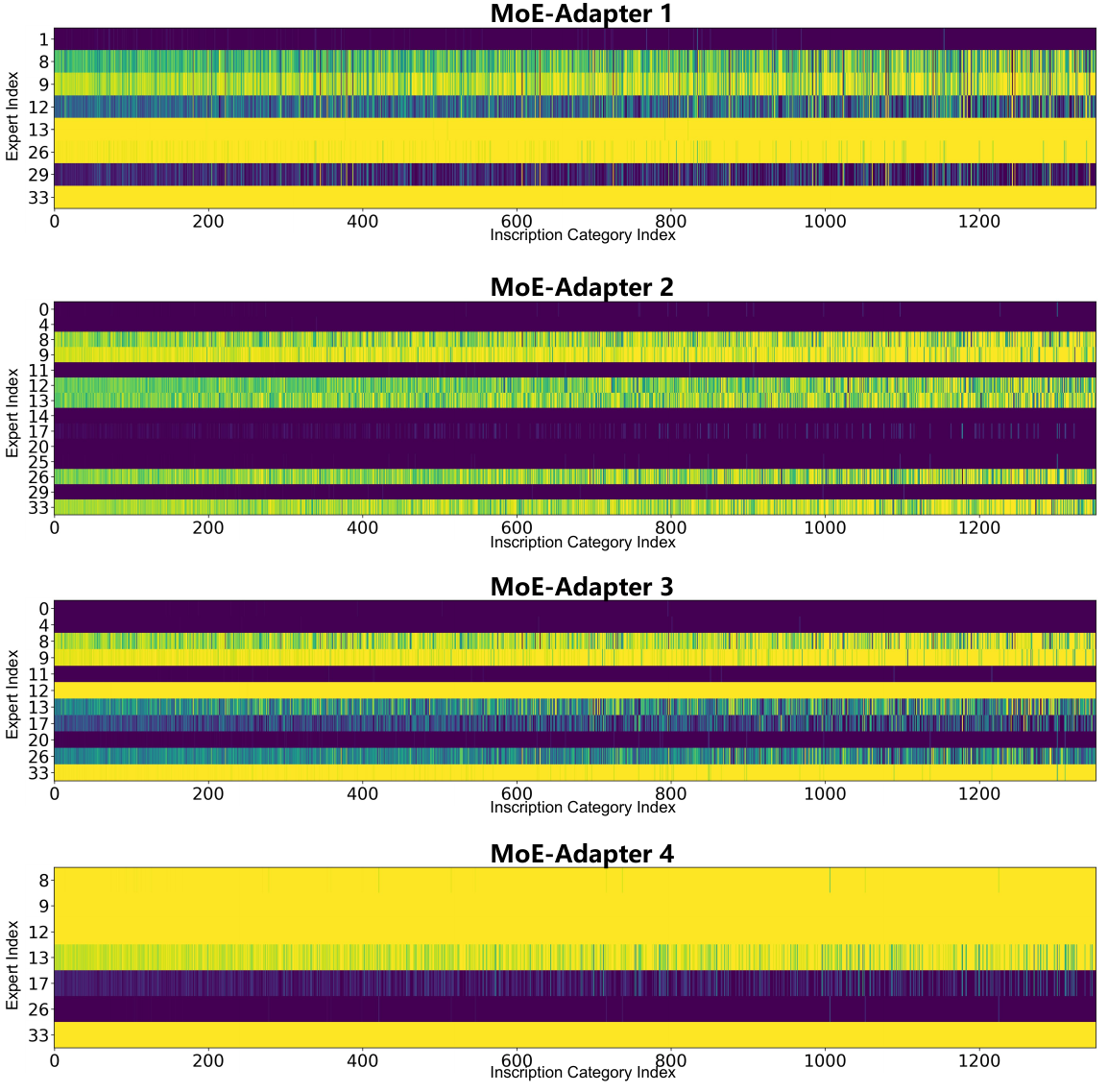}
  \caption{Visualization of expert activation frequencies across four MoE-Adapters in our Ladder Side framework. The horizontal axis represents character category indices, while the vertical axis denotes expert indices. Bright regions indicate higher activation frequency.} 
  \label{fig: expert analyse}
\end{figure*}

We perform a series of ablation studies to quantify the contribution of each key component in our framework by overall accuracy metric, as shown in Figure~\ref{fig: ablation}. \textit{(a) Number of Experts in MoE Adapters.} Disabling the MoE module (0 experts) yields an overall accuracy of 77.5\%. Accuracy dips near 76.5\% at 9 experts, then rises steadily to the best performance of 78.8\% with 36 experts. This monotonic upward trend after 9 experts indicates that enlarging the expert pool provides richer specialization and stronger representation learning. \textit{(b) Top-k Selection.} As the router’s top-k selection increases from k = 2 to k = 5, accuracy remains around 78.5\% for k = 2–4 but reaches 78.8\% at k = 5. The upward tendency suggests that allowing the router to activate a broader subset of experts facilitates more comprehensive feature aggregation. \textit{(c) Ordered Sequence Fine-tuning (OSF) Epochs.} Without OSF fine-tuning (0 epochs), the model attains only 77.93\% accuracy. Performance climbs with more OSF epochs, peaking near 78.8\% at 5 epochs, then shows mild oscillations with further training. This confirms that moderate OSF training effectively reinforces correct character order, while excessive fine-tuning brings no additional gain. \textit{(d) Permuted Sequence Number in PLM.} Increasing the number of permuted sequences from 4 to 12 improves accuracy from 78.1\% to 78.8\%, highlighting that richer permutation diversity strengthens sequence modeling.

Notably, the relationships in ablation studies (a), (b), and (d) all exhibit a generally ascending trend between parameter magnitude and performance. Although we observe consistent gains at the tested upper bounds (36 experts, top-5 routing, 12 permutations), resource constraints prevented exploration beyond these settings, leaving open the possibility of further improvements with larger configurations.

\subsection{Analysis of Experts Selection}

Figure~\ref{fig: expert analyse} shows the expert activation frequencies of four MoE-Adapters on the test set during inference. Within each adapter, the distribution of activated experts is highly non-uniform: only a small subset of experts are frequently selected, while the majority remain rarely utilized. 

When comparing across different adapters, one can observe both overlap and divergence. Certain expert indices (e.g., 9 and 33) are frequently selected in multiple MoE adapters, suggesting that these experts capture universally useful features across character categories. At the same time, different MoE adapters also activate some unique experts internally, indicating that they specialize in complementary subspaces. This inter-adapter diversity suggests that while individual adapters are prone to expert sparsity, the ensemble of multiple adapters ensures broader coverage of the expert pool, thereby enhancing the model’s representation capacity.

\section{Conclusion}\label{sec6}

We presented a large-scale bronze inscription (BI) dataset and a two-stage detection–recognition pipeline that first localizes inscriptions and then transcribes individual characters. To address the key challenges of cross-domain variability, visual degradation, and extreme class imbalance in BI recognition, we propose LadderMoE, a parameter-efficient recognizer that augments a pretrained CLIP encoder with ladder-style mixture-of-experts adapters for dynamic expert specialization. Comprehensive experiments on single-character and full-page tasks confirm that the integrated system consistently surpasses leading scene-text recognition baselines across head, mid, and tail categories and across color, rubbing, and tracing domains, offering a robust and scalable foundation for automatic bronze-inscription recognition and for downstream archaeological analyses.

\bibliography{sn-bibliography}% common bib file

\end{CJK*}

\end{document}